\documentclass[sigconf]{acmart}

\renewcommand\footnotetextcopyrightpermission[1]{} 
\usepackage{wrapfig,balance,colortbl}
\usepackage{amssymb}
\usepackage{latexsym}
\usepackage{amsmath}
\usepackage{alltt}
\usepackage{graphicx}
\usepackage{url}
\usepackage{amsfonts}
\usepackage{color}
\usepackage{float} 
\usepackage{multicol}
\usepackage{multirow}
\usepackage{hhline}
\usepackage{graphicx}
\usepackage{booktabs}
\usepackage{rotating}
\usepackage{caption}
\usepackage{algorithm}
\usepackage{algorithmicx}
\usepackage[caption=false,font=footnotesize]{subfig}
\usepackage{algpseudocode}
\usepackage{tabto}
\usepackage{lipsum}
\usepackage{pbox}
\usepackage{array}
\usepackage{enumitem}
\usepackage{xcolor}
\usepackage{acmcopyright}
\usepackage{array,multirow,graphicx}

\definecolor{Gray}{gray}{0.9}

\newcommand{\exclude}[1]{}

\include{psfig}

\usepackage[normalem]{ulem}

\begin{document}

\title{A Mixture of Expert Approach for \\Low-Cost Customization of Deep Neural Networks \vspace{-4mm}}

\vspace{-5mm}
\author{Boyu Zhang, Azadeh Davoodi, and Yu-Hen Hu}
\affiliation{%
  \institution{University of Wisconsin - Madison}
}
\email{{bzhang93, adavoodi, yhhu}@wisc.edu}
\vspace{-5mm}

\begin{abstract}
The ability to customize a trained Deep Neural Network (DNN) locally using user-specific data may greatly enhance user experiences, reduce development costs, and protect user's privacy. In this work, we propose to incorporate a novel Mixture of Experts (MOE) approach to accomplish this goal. This architecture comprises of a Global Expert (GE), a Local Expert (LE) and a Gating Network (GN). The GE is a trained DNN developed on a large training dataset representative of many potential users. After deployment on an embedded edge device, GE will be subject to customized, user-specific data (e.g., accent in speech) and its performance may suffer. This problem may be alleviated by training a local DNN (the local expert, LE) on a small size customized training data to correct the errors made by GE. A gating network then will be trained to determine whether an incoming data should be handled by GE or LE. Since the customized dataset is in general very small, the cost of training LE and GN would be much lower than that of re-training of GE. The training of LE and GN thus can be performed at local device, properly protecting the privacy of customized training data. In this work, we developed a prototype MOE architecture for handwritten alphanumeric character recognition task. We use EMNIST as the generic dataset, LeNet5 as GE, and handwritings of 10 users as the customized dataset. We show that with the LE and GN, the classification accuracy is significantly enhanced over the customized dataset with almost no degradation of accuracy over the generic dataset. In terms of energy and network size, the overhead of LE and GN is around 2.5\% compared to those of GE.
\vspace{-2mm}
\end{abstract}

\maketitle

\vspace{-2mm}
\section{Introduction}

The use of Deep Neural Network (DNN) allows efficient embodiment of intelligence into emerging application domains such as automotive, healthcare, etc \cite{deepdriving, deeplanguagetrans}. Often, an intelligent assistant such as Siri or Alexa will be equipped with sensory devices (cameras, microphones) to capture and transmit the sensory data to the cloud to be processed by a DNN in the backend for speech recognition or human face detection or other cognitive tasks. While a cloud-tethered model may provide acceptable performance for occasional use, it may be insufficient to provide required performance when presented with user-specific data (e.g., accent in speech). This architecture may also fail when a network connection is unavailable. When the cognitive task is to sense user's biometric data, sending data to cloud for processing may also raise privacy concern.

An alternative approach would be to leverage edge computing that utilizes local resources to facilitate DNN based inference without tethering to cloud. Several approaches \cite{quantization, lowrankexpansion, neuronelimination} have been proposed to approximate a trained DNN so that it may be implemented on a stand-alone embedded device which may contain DNN accelerators such as a TPU \cite{tpu}. However, such a DNN is often trained with large (millions) scale training dataset on a data center infrastructure. Modifying its structure or weight values cannot be easily accomplished on an edge device. Moreover, we envision that such trained (and down-sized) DNN may be provided by vendors in the form of intellectual property and cannot be modified by the user. Under these situations, it would be desirable to empower the edge computing devices with local learning capability so that the behavior of the intelligent assistants may be customized to correct mistakes made by the trained DNN and improve performance.

In this work, we propose a novel architecture and design methodology to customize a DNN that is deployed on an edge device with low-cost hardware overhead. This DNN has already been trained on a large generic dataset that is representative of many users. We assume that initially, the trained DNN provides a reasonable level of performance. However, as more situation-dependent, customized data are presented to it, mistakes may occur. When the DNN makes a mistake, the user can correct it (by over-writing the DNN's output). This user annotation (correction) of data then will be used for local learning with the goal of not to repeat the same mistake.

Our proposed approach for situation-aware, personalized, local learning is based on a Mixture of Expert (MOE) architecture. In the MOE architecture, the DNN trained with generic dataset is viewed as a Global Expert (GE). We then use a small DNN as a Local Expert (LE). The LE will be trained with a small size customized training data to correct the errors made by the GE. The third component of the MOE architecture is a Gating Network (GN), also represented by a small DNN, which determines whether an incoming data should be handled by the GE or LE. Since the customized dataset is very small, the overhead of implementing and training the additional components (i.e., LE and GN) will also be very small. Another distinguishing feature of the MOE architecture is the ability to completely disable either the general or the customization mode by means of power-gating the corresponding components. Moreover, the flexible architecture of the MOE may be easily extended to multiple local experts (LEs) to accommodate different classes of users, if desired.

In this work, we develop a prototype MOE architecture for the task of recognizing handwritten digits and characters. We use EMNIST \cite{emnist} as the generic dataset and representative of many users. We then use a small customized dataset provided by \cite{aspdac2018} which is representative handwritings of 10 users. In our experiments, we show the overhead of LE and GN is about 2.5\% of GE in terms of energy and the network size (represented by the number of distinct parameters). This is while achieving significant improvement in the classification accuracy over the testing set of the customized data (i.e., on average from 75.09\% to 92.58\%). Moreover, there is almost no degradation in the classification accuracy when the generic, non-customized data (i.e., the testing data in EMNIST) is used.

Compared to a recent work \cite{aspdac2018} which also focuses on customization, the MOE architecture provides more flexibility and results in achieving higher classification accuracy over the generic dataset (69.15\% vs 74.48\%). In terms of overhead in network size and energy consumption the MOE model is only 58.79\% and 58.28\% of \cite{aspdac2018}.
\enlargethispage{\baselineskip}

Overall, the summary of our contributions are as follows:
\begin{itemize}
    \item We propose the use of MOE architecture as a low-cost hardware solution to customize a trained DNN, which represents a general model for many users.
    \item We propose novel techniques based on \textit{structure sharing} which are specifically designed to reduce the hardware cost to implement the LE and GN components.
    \item We show the utility of the MOE architecture by developing a prototype implementation for recognizing customized handwritten digits and characters.
\end{itemize}
We first discuss design methodology and architecture of the MOE in Section \ref{sec:moe}. Comparison with related work is in Section \ref{sec:related}. Simulation results are discussed in Section \ref{sec:experiments} followed by conclusion.






\begin{figure}[t]
        \centering
        \includegraphics[width=3.4in]{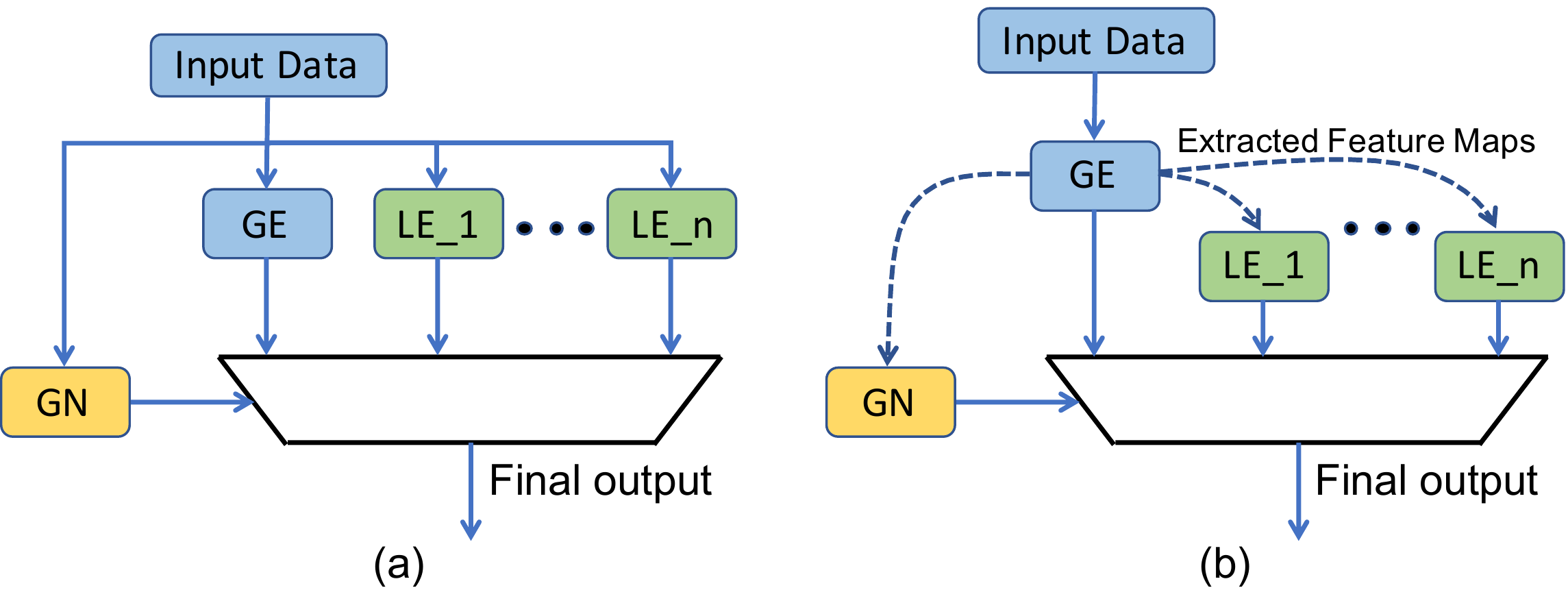}\vspace{-5mm}
        \caption{Diagram of the MOE architecture before (a) and after (b) structural optimization}\vspace{-4mm}
        \label{fig:moe}
\end{figure}
\vspace{-2mm}
\section{Design Methodology and Architecture of MOE}\label{sec:moe}

A block diagram of the proposed MOE architecture is shown in Fig. \ref{fig:moe}(a). It consists of a global expert (GE), one or more local experts (LEs), and a 
gating network (GN) that provides a data dependent \textcolor{black}{selection signal to select the output of either} the GE or one of the LEs. In the case of multiple LEs, each one may represent one class of users. Ideally, if the GE provides a correct output with respect to a given input, the GN will just pass through GE's output as the final output and suppress LE's output. If an LE is more likely to provide a correct output, then the output of the GE (and other LEs) will be suppressed. Therefore, the GN plays a central role in the MOE architecture. For each input feature vector, it predicts which of the experts (GE and LEs) is more likely to provide a correct output and then connect the output of that expert to the final output. In this manner, the GN behaves like an intelligent multiplexer, multiplexing the outputs of GE and LEs according to the predicted likelihoods of their output to be correct. For simplicity, we only discuss the case with one LE in the subsequent sections but the discussions can be easily extended to more than one LE.


\vspace{-2mm}
\subsection{Details of the Three Components in MOE}\label{sec:threecomponents}

\noindent \textbf{Global Expert:} We envision there will be an IP (intellectual property) market for trained DNN on very large generic dataset. These carefully designed and trained DNNs may be licensed by vendors to be incorporated into intelligent cognitive devices to facilitate speech recognition, face recognition, and other intelligent services. These trained DNNs will be designated as the Global Expert (GE). The weight values and internal structures are proprietary information and cannot be revealed to or modified (re-trained) by individual users. Therefore, given an input feature vector, a GE will provide the corresponding output, and corresponding internal neural outputs (e.g., extracted features) to the user, but cannot be retrained at the embedded device. Since the GE is trained with generic dataset, it is expected to make a few mistakes on customized data. The users may then opt to provide these mistakenly classified customized data back to the vendor. As such, the vendor may further design and train the GE to improve its performance. Then the updated GE is released to users as a firmware update. However, such an update is expected to be a rare event and can be disabled if the communication with cloud is unavailable. 



\noindent \textbf{Local Expert:} A Local Expert (LE) is a local trainable DNN residing within the intelligent cognitive device. It will be trained on a small size local customized dataset with the purpose to correct the errors made by the GE. The customized training dataset is collected by the system during daily uses.
Since the size of the customized dataset is much smaller than that of the generic dataset, LE would have a much smaller and simpler DNN structure compared to that of the GE. This lightweight structure incurs less hardware overhead, which is critical for deploying the entire system on embedded and mobile platforms. Moreover, because both the customized dataset and the LE are small, the training of LE can be easily carried out by software or co-processor on-chip.


\noindent \textbf{Gating Network:}  In its most generic form, the gating network (GN) can be viewed to have two outputs: $w_G$, and $w_L$ such that $0\leq w_G$, $w_L\leq 1$ and $w_G + w_L = 1$. The outputs of GE and LE will be weighted by $w_G$, and $w_L$ respectively; $w_G$ will be multiplied to each output of the GE and $w_L$ will be multiplied to each output of the LE. The results of these multiplications then will be added together to form the final output for each class. The GN acts as a mediator between the GE and the LE: 
If it is likely that GE will provide the correct answer, then $w_G >> w_L$, meaning the GE's output will be selected as the final output. On the other hand, if it is more likely that the LE's output is correct, then $w_G << w_L$ meaning the LE's output will be the final output.   

In this work, we use a specific realization of the above GN description. First, GN is implemented as a DNN with two outputs corresponding to the likelihoods of selecting GE and LE, respectively. Based on these two outputs, the weights $w_G$ and $w_L$ are then generated as binary values so GE is selected if the DNN output corresponding to GE has a higher likelihood than the one corresponding to LE, and vice versa.  Therefore, in our specific realization, the GN's task is to strictly select between GE or LE in order to generate the final output. The DNN of GN is trained with a mixture of the customized dataset used for training LE and equal number of generic dataset (randomly sampled from the dataset used for training the GE). Since the customized dataset is much smaller than that of the generic dataset, only about 1\% of data from the generic dataset is sampled. The customized data samples share a common label of [0,  1] ($w_G = 0, w_L = 1$), and the sampled generic data samples have a common label [1,  0] ($w_G = 1, w_L = 0$).

\vspace{-3mm}
\subsection{Optimization of LE and GN Architecture}

As can be seen in Fig. \ref{fig:moe}(a), all three main components of the MOE architecture, GE, LE, and GN share the same input feature vectors. The GE, being a licensed IP of a trained DNN, will provide its output. However it can also provide some extracted features from GE, to both the LE and the GN as shown in Fig. \ref{fig:moe}(b). We assume such a structural sharing mechanism is available which can be incorporated into the license agreement perhaps at additional costs. 

Structural sharing can help provide a very good starting point to LE and GN because GE is trained with large size generic dataset and the customized data share the same fundamental properties as the generic ones. More importantly, it can be leveraged to simplify the structural designs of DNNs for both the LE and the GN.

More specifically, instead of feeding raw input data to the DNNs of LE and GN, we have these DNNs to draw intermediate feature maps from the internal outputs (outputs of hidden neurons) of the GE.  
Thus the GE shares one or more of its convolutional layers (for higher level feature extraction) with both the LE and GN. This is done by providing \textit{only} the intermediate (higher level) features extracted from the input without revealing GE's internal weight values. Thus, the only information revealed (to a hacker) would be the dimension of these intermediate feature maps.  


Sharing of feature maps effectively helps eliminate the convolutional layers from LE and GN and simplify the structures of these two DNNs to only fully-connected layers, as we show in our experiments. Despite the above improvement, the size of the extracted features from GE may still be large, which makes the fully connected layer in LE or GN to also become large. For example, as we discuss in our experiments, the size of the extracted convolutional layer in GE is $12\times 12\times 20$ which results in a large fully-connected layer in LE and GN. In a more optimized architecture, the LE and GN only \textit{partially} use the intermediate features from the GE. One way to achieve this is to use a max pooling layer to further reduce the width and height of the feature maps from GE, and then feed the sub-sampled feature maps to the subsequent fully connected layer in LE and GN. 
This approach treats each feature map equally and is easy to implement. Note, in general the LE and GN may choose different degrees to sub-sample the feature maps from GE. (We will provide the specifics about how this max pooling is implemented using the prototype case study in Section 4.)





\vspace{-3mm}
\subsection{An Alternative Training Option}\vspace{-1mm}


A pattern classifier such as the GE or the LE will partition the customized feature space $R$ into two disjoint sub-regions: one in which the classifier's output is deemed correct and the other incorrect, over the customized data. 
In the table below, let us consider four disjoint regions $\{R_k; 1 \le k \le 4 \}$ where the outputs of the GE and the LE will be correct (Y) or incorrect (N). The row titled with GN gives the desired output of the GN where the lower-case letter d represents either 1 or 0, a don't care situation. 

\begin{table}[h]
\centering \vspace{-4mm}
\begin{tabular}{c|cccc}
\multicolumn{1}{l}{} & \textbf{$R_1$} & \textbf{$R_2$} & \textbf{$R_3$} & \textbf{$R_4$} \\
\midrule
\textbf{GE}          & Y           & N           & Y           & N           \\
\textbf{LE}          & Y           & Y           & N           & N           \\
\textbf{GN}          & {[}d d{]}   & {[}0 1{]}   & {[}1 0{]}   & {[}d d{]}   \\
\textbf{MOE}         & Y           & Y           & Y           & N          
\end{tabular}\vspace{-2mm}
\end{table}
Initially, given the training data set $R$, the GE will partition it into $R_1\cup R_3$ (correct classification) and $R_2\cup R_4$ (mis-classification). The LE will be trained on $R$ and partition it into $R_1\cup  R_2$ (correct classification) and $R3\cup  R4$ (mis-classification). From these results, the four regions $R_1$, $R_2$, $R_3$, and $R_4$ can be identified.

From this table, using MOE, the classification accuracy may be increased from $R_1\cup R_3$ using only GE to up to $R_1\cup R_2 \cup R_3$ using GE and LE. To achieve this performance enhancement, the local expert LE should strive to correct mistakes made by the GE though by maximizing region $R_2$ and shrinking region $R_4$.

\begin{figure}[t]
        \centering
        \includegraphics[width=2.1in]{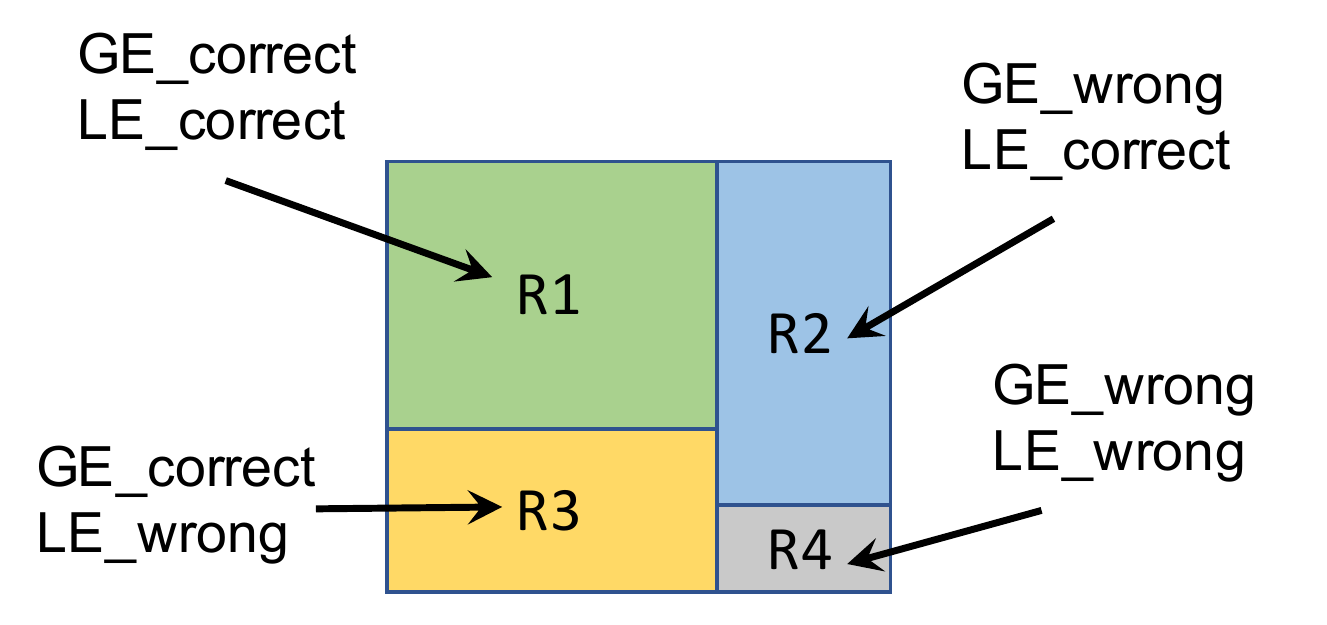}\vspace{-3mm}
        \caption{Diagram of the sample region covered by GE and LE. Region $R1+R3$ represents the samples that are already  correctly classified by GE. Region $R2$ represents the samples that can \textit{only} be correctly classified by LE. }\vspace{-5mm}
        \label{fig:set}
        \vspace{-3mm}
\end{figure}

This is also pictorially shown in Fig. \ref{fig:set}. The outermost rectangle ($R_1 \cup R_2 \cup R_3 \cup R_4$) represents all customized samples. $R_1 \cup R_3$ represents the samples that can be correctly classified by GE. $R_1 \cup R_2$ represents the samples that can be correctly classified by LE. $R_4$ represents the samples that cannot be correctly identified by  GE and LE. The ability of LE to maximally complement GE is shown by the relative size between $R_2$ and $R_4$. If $R_2$ is much larger than $R_4$, it indicates that LE is able to provide correct predictions on those samples that GE cannot handle, and LE complements GE very well. 


The above requires to train LE with samples in $R_2\cup R_4$, and train GN to identify $R_2\cup R_4$ from the rest of the region $R$. This is equivalent to changing the functionalities of LE and GN to identify when GE makes a mistake (regardless of handling customized versus generic data).
However, our training experiences showed overfitting issue may occur due to insufficient training samples in $R_2\cup R_4$; for example the size of $R_2\cup R_4$ is only about 25\% of $R$ and $R$ is already very small in our experiments. 
Thus, in this work, we train LE on all customized data and train GN to predict if an input data is customized or generic, as explained in Section 2.1.  LE still complements GE very well under this training setting. 

\vspace{-3mm}
\section{Related Work}\label{sec:related}\vspace{-1mm}

The idea of Mixture of Experts was originally proposed in the 90s \cite{moehinton, moejacob}. For example, it was shown to be very effective for patient-adaptable electrocardiogram beats classification \cite{moehu}. However, the problems that were considered at that time were considerably less complex and resulted in significantly simpler models compared to the ones today \cite{moegoogle}. Furthermore, with increasing emphasis on edge computing, power efficiency becomes a critical factor during the design of the components of MOE. Thus, unlike previous works that focused on different expert architectures \cite{moesvm, moegaussian} and configurations \cite{moeinfinite}, our work focuses on how to integrate the general MOE concept with DNN to achieve higher performance on complex tasks with minimal hardware overhead.

Along the same line of customizing DNN for different users, a recent prior work \cite{aspdac2018} introduced embedding the base DNN with a task-specific network and an aggregation layer. Compared with their architecture, the proposed MOE architecture has more flexibility due to the use of GN instead of an aggregation layer. The MOE architecture also has higher interpretability level because the output of each component can be easily extracted and analyzed, whereas the behaviors of the task-specific network and the aggregation layer are hard to analyze. Moreover, with realization of a high performance GN, the MOE architecture will have negligible performance degradation on generic dataset as we show in our experiments. 

In fact, the customization of both our work and \cite{aspdac2018} can be viewed as containing elements from transfer learning. The general idea of transfer learning is to improve the learning of target task on target domain by utilizing the knowledge learned from source domain regarding to source task \cite{transferlearning}. In our setting, the target task and the source task are the same, which is classifying handwritten alphanumeric characters. The source domain is composed of generic dataset, and the target domain consists of customized dataset. In the design of both LE and GN, we transfer the knowledge learned by the convolutional layers in GE and then perform fine-tuning \cite{finetuning} and joint-training \cite{jointtraining} on LE and GN, respectively. The reason we choose to share the outputs of convolutional layer of GE rather than fully connected layer is because the features extracted by the first layer(s) of a DNN are more general \cite{transferablefeature} and we believe these features are indeed shared by generic and customized datasets. The extracted generic features are then processed by more subsequent layer(s) in both LE and GN to make their own specific decisions. Another way of performing transfer learning is by fine-tuning the last few layers of a DNN. We compare the results of this approach and the proposed MOE model in Section \ref{sec:compare}. \vspace{-4mm}

\section{Experimental Setup and Results}\label{sec:experiments}


\subsection{Generic and Customized Datasets}
\label{sec:dataset-info}

We evaluate the proposed MOE model on the classification problem of recognizing handwritten digits and letters. The Extended MNIST (EMNIST) dataset \cite{emnist} is used as the generic dataset during training. The images of handwritten digits and letters released by \cite{aspdac2018} are used as the customized dataset for customized training and testing.

\noindent \textbf{Generic Dataset:} The EMNIST dataset is generated by applying Gaussian blurring, centering, padding, and down sampling to all the images in NIST Special dataset 19 \cite{nist}. The entire dataset is released by \cite{emnist}. The structure of the EMNIST dataset is exactly the same as that of the NIST dataset. Specifically, the EMNIST dataset contains 814,255 $28\times 28$ grayscale images in total, and each of these images belongs to one of 62 classes: digits `0'-`9', lower-case letters `a'-`z', and upper-case letters `A'-`Z'. These images are further divided into training and testing sets with the same probability for each class, which results in 697,932 and 116,323 images in training and testing sets, respectively. Since the number of samples in each class is highly unbalanced in both training and testing sets, we first balanced the classes in these two sets before using them during training. Specifically, for both sets, we identified the class that has the least number of samples in each set and then randomly down sample other classes so that the number of samples in other classes roughly matches that of the class we identified at the beginning in the corresponding set. This results in about 2700 samples per class in training set and about 453 samples per class in testing set. Overall, there are 165,092 samples in training set and 27,537 samples in testing set, which correspond to 23.65\% and 23.67\% of the original training and testing sets.

\noindent \textbf{User Customized Dataset:} The user customized dataset is collected and released in \cite{aspdac2018}. It contains $28 \times 28$ grayscale images of 62 alphanumeric characters written by 10 users. All images are collected and pre-processed using similar techniques as those in the EMNIST dataset. For each user, it has 30 images per class in training set and 10 images per class in testing set. This results in a total of 1860 training and 620 testing images per user. Compared with the size of the generic dataset, this corresponds to 1.13\% and 2.25\% in terms of training and testing sets, respectively.
\vspace{-3mm}
\subsection{The MOE Network Structure}

\begin{figure}[t]
       \centering
       \includegraphics[width=1.8in]{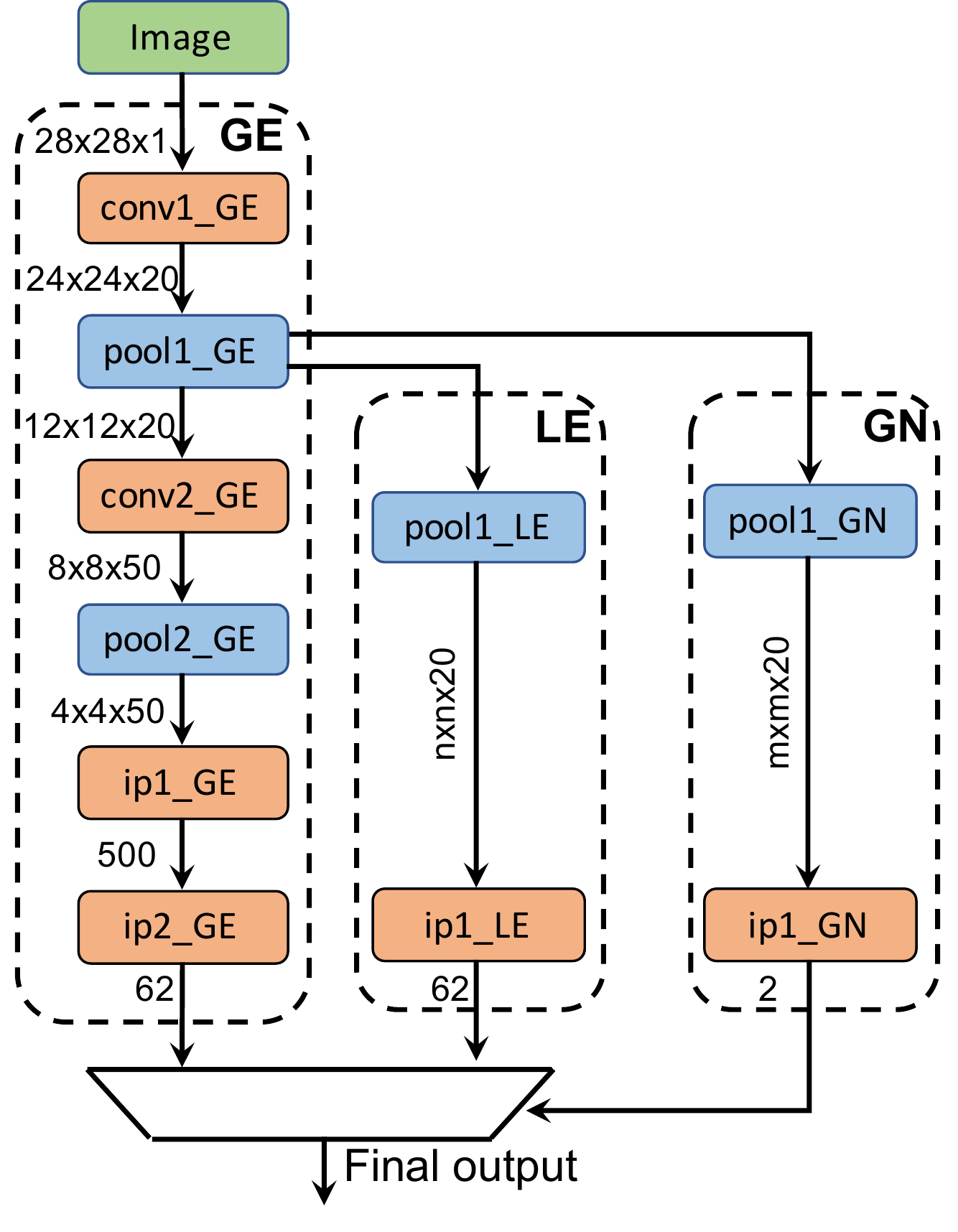}\vspace{-6mm}
       \caption{Network structure of the prototype MOE model for the customized EMNIST recognition task}\vspace{-8mm}
       \label{fig:network}
\end{figure}

Fig. \ref{fig:network} shows the network structure of the prototype MOE model used in the experimented classification task. The GE is a variation of LeNet5 \cite{lenet5} that is modified to produce 62 outputs. The LE shares its first convolutional layer with the GE by using the output feature maps of the first pooling layer of GE as its input feature maps. The LE then performs further max pooling operation on its input feature maps, which down samples them to the size of $n \times n \times 20$. The exact value of $n$ depends on the receptive window size and stride of the kernel of the pooling layer. We will discuss how the value of $n$ is determined in our experiments to achieve the minimum overheads in hardware implementation cost with negligible loss in classification accuracy. The down-sampled feature maps is then fed into a fully connected layer that outputs the result of LE's classification. The size of this fully connected layer is $n\times n\times 20\times 62$.

The GN has a similar structure to the LE with the exceptions that: 1) the size of the down-sampled feature maps is $m\times m\times 20$ so it can be different from the size of LE; 2) its fully connected layer only outputs two values indicating whether the final classification result should be selected from GE or LE, therefore the size of this layer is $m\times m\times 20\times 2$. Similar to the LE case, we will discuss how the value of $m$ is determined in our experiments.

\vspace{-3mm}
\subsection{Training Procedure}
\label{sec:training-procedure}

Tensorflow  is used for building and training the entire network. The training is performed in three steps as described below:
\begin{enumerate}
   \item The GE is trained with the training set from the generic dataset as described in Section \ref{sec:dataset-info}. Note, in practice, this training is done prior to deployment on-chip. After training, all the parameters in GE are fixed in the subsequent steps.

   \item For each user, only the fully connected layer in LE is trained with the training set from that user's customized dataset as described in Section \ref{sec:dataset-info}.

   \item For each user, only the fully connected layer in GN is trained with the training set from that user's customized dataset plus equal number of samples from the generic dataset. The label of each training sample is either [1, 0] or [0, 1] indicating whether the sample is from generic  or customized dataset.
\end{enumerate}

The first step is referred as generic training and the last two  are referred as customized training. During all aforementioned training steps, a small portion of the corresponding training set is used as validation set to tune the hyper-parameters of the training process, such as learning rate and number of training epoch.

\vspace{-3mm}
\subsection{Results}

\subsubsection{Determining the size of pooling layers in LE and GN} We experimented with different values of $m$ and $n$ to determine the sizes of the pooling layers in LE and GN which are denoted by pool1\_LE and pool1\_GN in Fig. \ref{fig:network}. Specifically, for each set of values of $m$ and $n$, we trained the LE and GN networks and recorded classification accuracy as well as the hardware overhead of the size of LE and GN, relative to GE. Hardware overhead was measured with respect to number of distinct parameters in LE and GN.

Table \ref{tab:exploration} reports the percentage increase in network size due to LE and GN relative to GE, and the classification accuracy of the overall network which is reported as separate quantities over the customized and generic testing datasets. For each row, the same value is used for $m$ and $n$ so the sizes of the pooling layers in GN and LE are equal to each other in our implementation. Column 1 reports the size of the down-sampled input feature maps in LE and GN. The first row shows the default size if pooling is not performed which as can be seen results in over 40\% overhead in network size.

As can be seen the overhead in network size drops dramatically from 40.38\% to 0.28\% as the down-sampled input feature maps becomes smaller and smaller. The accuracies on both customized data and generic data remain relatively stable before the down-sampled input feature maps size shrinks to $2 \times 2 \times 20$ (so $n=m=2$). After that, both accuracies experience a noticeable drop. The highlighted row represents a good configuration of LE and GN in the sense that it only incurs minimal network size overhead while still provides very high accuracies. Thus, it is used in the following experiments.

\begin{table}[t]
\scriptsize
\tabcolsep=2.9pt

\centering
\caption{Overhead of network size and overall classification accuracy for different degrees of subsampling from GE}\vspace{-3mm}
\label{tab:exploration}
\begin{tabular}{c|c|cc}
\toprule
\multirow{2}{*}{\textbf{\begin{tabular}[c]{@{}c@{}} Input Feature\\Map Size in LE/GN\end{tabular}}}
& \textbf{Network Size (\%)}
& \multicolumn{2}{c}{\textbf{Overall Accuracy}} \\
& $|LE+GN|/|GE|$ & Customized & EMNIST                 \\
\midrule
12$\times$12$\times$20    & 40.38\%    & 91.58      & 75.57     \\
6$\times$6$\times$20      & 10.09\%    & 92.53      & 75.52     \\
4$\times$4$\times$20      & 4.49\%     & 92.81      & 75.24     \\
\rowcolor{Gray}
3$\times$3$\times$20      & 2.52\%     & 92.58      & 74.48     \\
2$\times$2$\times$20      & 1.12\%     & 89.03      & 72.29     \\
1$\times$1$\times$20      & 0.28\%     & 47.77      & 69.33     \\
\bottomrule
\end{tabular}\vspace{-5.5mm}
\end{table}

\vspace{-2mm}
\subsubsection{Classification accuracies of various components in  MOE}
Table \ref{tab:acc-various} reports the performance of various components of the MOE model after the training is complete. For the customized data:
\begin{itemize}
   \item Column `GE' measures the accuracy of GE on customized testing set. It represents the case that only the GE is available and provides a reference point for evaluating the performance of the proposed MOE model.

   \item Column `LE' measures the accuracy of LE on the customized testing set. It evaluates how well the LE performs in classifying user customized data \textit{if} the GN behaves perfectly.

   \item Column `GN' measures the accuracy of GN on customized testing set. It evaluates how well GN performs in classifying if the input image is from customized vs generic datasets.

   \item The `Overall' column measures the final accuracy of the proposed MOE model on the customized testing set.

   \item Column `LE|GE$_{wrong}$' measures the accuracy of LE on the customized testing samples that are wrongly classified by GE. It is a metric reflecting how well LE complements GE.
\end{itemize}
For generic data, the last column of Table \ref{tab:acc-various} reports the accuracy of the entire model on the testing set of the generic dataset. The results are reported for 10 users in the customized datasets.

As it can be seen from Table \ref{tab:acc-various}, on average, the accuracy of GE is only 75.09\% on the customized dataset, while LE is able to provide 92.82\% accuracy after the customized training (assuming GN behaves perfectly).  Fortunately, after customized training, GN performs very well with 97.95\% accuracy in distinguishing if an image is from customized dataset or generic dataset. Thus, the overall performance of the entire model is also very good with 92.58\% accuracy on average.

The table also suggests that LE is indeed a good complement for GE on customized data, because 84.85\% of time, the LE is able to provide correct prediction on those samples that cannot be correctly predicted by GE. We also note that the accuracy of the proposed MOE model on generic dataset after customized training is 74.48\%, indicating a minimal drop compared to that of the GE itself, which is 75.72\% as tested. (This accuracy of GE for the generic dataset is not reported in the table.)

\begin{table}[t]
\centering
\scriptsize
\tabcolsep=4.5pt
\caption{Post-training accuracy of various components in the MOE Model}\vspace{-3mm}
\label{tab:acc-various}
\begin{tabular}{c|ccccc|c}
\toprule
\multirow{2}{*}{\textbf{\begin{tabular}[c]{@{}c@{}}User\\ID\end{tabular}}}
& \multicolumn{5}{c|}{\textbf{Customized Data}}
& \textbf{EMNIST} \\
& GE   & LE  & GN & Overall & LE|GE$_{wrong}$ & Overall       \\
\midrule
\textbf{1}      & 65.16       & 86.29      & 99.35  & 86.45   & 72.69      & 74.96         \\
\textbf{2}      & 79.84       & 97.26      & 98.55  & 97.10   & 92.80      & 74.55         \\
\textbf{3}      & 78.71       & 95.00      & 96.77  & 94.84   & 87.12      & 74.77         \\
\textbf{4}      & 80.97       & 96.13      & 98.06  & 95.81   & 85.59      & 74.13         \\
\textbf{5}      & 70.48       & 90.65      & 98.71  & 90.48   & 80.87      & 74.74         \\
\textbf{6}      & 81.61       & 91.13      & 94.03  & 90.81   & 76.32      & 72.40         \\
\textbf{7}      & 71.24       & 93.38      & 98.55  & 93.05   & 87.08      & 74.22         \\
\textbf{8}      & 75.97       & 93.23      & 99.52  & 92.90   & 87.25      & 75.04         \\
\textbf{9}      & 79.35       & 97.26      & 98.87  & 97.26   & 93.75      & 74.81         \\
\textbf{10}     & 67.58       & 87.74      & 97.10  & 87.10   & 85.07      & 75.15         \\
\midrule
\textbf{Avg.} & 75.09       & 92.81      & 97.95  & 92.58   & 84.85      & 74.48         \\
\bottomrule
\end{tabular}\vspace{-3mm}
\end{table}

\begin{table}[t]
\scriptsize
\centering
\caption{Overheads in network size and energy consumption  of the LE and GN relative to GE}\vspace{-3mm}
\label{tab:overhead}
\begin{tabular}{c|cccc}
\toprule
\textbf{Network Size (\%)} & \multicolumn{4}{c}{\textbf{Overhead in Energy Consumption (\%)}} \\
$|LE+GN|/|GE|$   & MAC    & SRAM   & DRAM   & Total  \\
\midrule
2.52\%        & 0.50\%   & 2.58\%   & 2.52\%   & 2.45\%  \\
\bottomrule
\end{tabular}\vspace{-5mm}
\end{table}

Table \ref{tab:overhead} shows the overhead of network size and energy consumption of LE + GN, reported as a percentage relative to GE. For network size, we count the number of distinct parameters in both LE and GN and compare to that of GE. As for energy estimation, we use the energy model in \cite{ourpaper} and report energy break down of different components such as MAC, SRAM, and DRAM, and an overall overhead. Table \ref{tab:overhead} indicates that the overhead of network size and energy brought by LE + GN are minimal, at 2.52\% and 2.45\% respectively, relative to GE. 

\subsubsection{Comparison with \cite{aspdac2018} and fine-tuning the last few layers of pre-trained DNN} \label{sec:compare}

Besides the proposed MOE model, there exists other alternative approaches for user customization of pre-trained large DNNs. Specifically, the recent work \cite{aspdac2018} proposed to augment the pre-trained DNN with a task-specific network and an aggregation layer. Another approach that is widely used in transfer learning is to fix the parameters of the first few layers of the pre-trained DNN and fine-tune the last few fully connected layers by retraining them. In both models, the entire network (for \cite{aspdac2018}, including the task-specific network and aggregation layer) is first trained with generic data by service provider before shipping to customer.

Later on, after shipping to customer, only the task-specific network and the aggregation layer (for \cite{aspdac2018}), or the last few fully connected layers (for fine-tuning) are trained with customized data by user. In order to compare with these alternative approaches, we implemented these models in Tensorflow and carried out the training procedure as described in \cite{aspdac2018} with the same datasets mentioned in Section \ref{sec:dataset-info}. We also note that the same LeNet5 model is used as the pre-trained DNN as in the MOE model, and all of the two fully connected layers are retrained in the fine-tuning model since it generates better results than only retraining the last fully connected layer. The comparison of accuracy between these models and the proposed MOE model are shown in Tables \ref{tab:before-after-acc} and  \ref{tab:acc-comp}. We also compare network size and energy consumption between them.

Table \ref{tab:before-after-acc} shows the classification accuracies for the before and after customized training cases. Accuracies are reported separately over the customized and the EMNIST (generic) datasets. Compared to both \cite{aspdac2018} and fine-tuning, the MOE model has much lower accuracies on both datasets before customized training. This is because only the GE is trained with generic dataset and both LE and GN are initialized with random weights at this point in the MOE. Thus, LE and GN only introduce random noise to the entire network before customized training.

After customized training, the accuracies of the MOE model are dramatically increased to 92.58\% and 74.48\% as can be seen in both Table \ref{tab:before-after-acc} and Table \ref{tab:acc-comp}. Specifically, the accuracy of MOE model on customized dataset reaches to a level similar to that of \cite{aspdac2018} and fine-tuning. Meanwhile, the accuracy of the MOE model on generic data is 74.48\%, which is higher than \cite{aspdac2018}'s 69.15\% and fine-tuning's 71.96\%.  Compared to GE itself, this represents only 1.24\% accuracy degradation versus 6.57\% in \cite{aspdac2018} and 3.76\% in fine-tuning.

In terms of the overhead, we note that the network size and energy consumption overhead of the MOE model are only 58.79\% and 58.28\% of those of \cite{aspdac2018}. Energy consumption was measured similar to the previous experiment reported in Table \ref{tab:overhead}. Note that, the overheads of the MOE model relative to the fine-tuning approach has been effectively shown in \ref{tab:overhead}.

\vspace{-2mm}

\begin{table}[t]
\scriptsize
\centering
\caption{Comparison of accuracy of the overall network pre- \& post- customized training between the MOE Model and \cite{aspdac2018}}\vspace{-3mm}
\label{tab:before-after-acc}
\begin{tabular}{c|cc|cc}
\toprule
\multirow{2}{*}{\textbf{\begin{tabular}[c]{@{}c@{}}Model\\ Name\end{tabular}}} & \multicolumn{2}{c|}{\textbf{Before Training}} & \multicolumn{2}{c}{\textbf{After Training}} \\
& Customized     & EMNIST    & Customized    & EMNIST                \\
\midrule
\textbf{MOE}      & 46.35   & 45.62     & 92.58     & 74.48               \\
\textbf{\cite{aspdac2018}}   & 75.05   & 75.94     & 93.25     & 69.15    \\
\textbf{Fine-tuning}  & 75.09  & 75.72   & 93.36    & 71.96               \\
\bottomrule
\end{tabular}\vspace{-6mm}
\end{table}

\begin{table}[h]
\scriptsize
\centering
\caption{Comparison of post-training accuracy between the MOE model and \cite{aspdac2018}}\vspace{-3mm}
\label{tab:acc-comp}
\begin{tabular}{c|cc|cc|cc}
\toprule
 \multirow{2}{*}{\textbf{User ID}}
& \multicolumn{2}{c|}{\textbf{MOE}}
& \multicolumn{2}{c|}{\textbf{\cite{aspdac2018}}}
& \multicolumn{2}{c}{\textbf{Fine-tuning}} \\
                & Customized       & EMNIST           & Customized        & EMNIST          & Customized        & EMNIST  \\
\midrule
\textbf{1}      & 86.45            & 74.96          & 87.58             & 67.65             & 86.29             & 71.87\\
\textbf{2}      & 97.10            & 74.55          & 97.90             & 70.39             & 96.61             & 72.41\\
\textbf{3}      & 94.84            & 74.77          & 94.68             & 69.68             & 95.48             & 71.91\\
\textbf{4}      & 95.81            & 74.13          & 96.29             & 70.15             & 96.77             & 72.32\\
\textbf{5}      & 90.48            & 74.74          & 90.97             & 67.51             & 91.13             & 71.51\\
\textbf{6}      & 90.81            & 72.40          & 91.13             & 71.33             & 92.10             & 73.59\\
\textbf{7}      & 93.05            & 74.22          & 92.29             & 67.49             & 93.91             & 70.82\\
\textbf{8}      & 92.90            & 75.04          & 94.52             & 69.70             & 94.68             & 71.74\\
\textbf{9}      & 97.26            & 74.81          & 98.87             & 69.61             & 97.90             & 72.14\\
\textbf{10}     & 87.10            & 75.15          & 88.23             & 67.97             & 88.71             & 71.31\\
\midrule
\textbf{Average} & 92.58            & 74.48          & 93.25             & 69.15            & 93.36             & 71.96\\
\bottomrule
\end{tabular}\vspace{-5mm}
\end{table}

\section{conclusions}\label{sec:conclusions}
We proposed to incorporate a Mixture of Experts model to achieve user customization of a trained DNN on embedded edge devices. We discussed the design methodology and architecture of the various components of MOE model in order to achieve low-cost hardware implementation, such that localized training becomes feasible while achieving high classification accuracy. We showed the effectiveness of our work using a prototype MOE architecture for identifying handwritten characters and digits.
\vspace{-2mm}

\scriptsize
\bibliographystyle{abbrv}

\begin{thebibliography}{10}

\bibitem{jointtraining}
R.~Caruana.
\newblock Multitask learning.
\newblock In {\em Learning to Learn}, pages 95--133. 1998.

\bibitem{deepdriving}
C.~Chen, A.~Seff, A.~Kornhauser, and J.~Xiao.
\newblock Deepdriving: Learning affordance for direct perception in autonomous
  driving.
\newblock In {\em IEEE International Conference on Computer Vision}, pages
  2722--2730, 2015.

\bibitem{emnist}
G.~Cohen, S.~Afshar, J.~Tapson, and A.~van Schaik.
\newblock {EMNIST}: an extension of {MNIST} to handwritten letters.
\newblock {\em arXiv preprint arXiv:1702.05373}, 2017.

\bibitem{moesvm}
R.~Collobert, S.~Bengio, and Y.~Bengio.
\newblock A parallel mixture of svms for very large scale problems.
\newblock In {\em Advances in Neural Information Processing Systems}, pages
  633--640, 2002.

\bibitem{finetuning}
R.~Girshick, J.~Donahue, T.~Darrell, and J.~Malik.
\newblock Rich feature hierarchies for accurate object detection and semantic
  segmentation.
\newblock In {\em IEEE Conference on Computer Vision and Pattern Recognition},
  pages 580--587, 2014.

\bibitem{nist}
P.~J. Grother.
\newblock {NIST} special database 19 handprinted forms and characters database.
\newblock {\em National Institute of Standards and Technology}, 1995.

\bibitem{aspdac2018}
B.~Harris, M.~S. Moghaddam, D.~Kang, I.~Bae, E.~Kim, H.~Min, H.~Cho, S.~Kim,
  et~al.
\newblock Architectures and algorithms for user customization of {CNNs}.
\newblock In {\em IEEE Asia and South Pacific Design Automation Conference},
  pages 540--547, 2018.

\bibitem{quantization}
S.~Hashemi, N.~Anthony, H.~Tann, R.~I. Bahar, and S.~Reda.
\newblock Understanding the impact of precision quantization on the accuracy
  and energy of neural networks.
\newblock In {\em Automation \& Test in Europe Conference \& Exhibition}, pages
  1474--1479, 2017.

\bibitem{moehu}
Y.~H. Hu, S.~Palreddy, and W.~J. Tompkins.
\newblock A patient-adaptable ecg beat classifier using a mixture of experts
  approach.
\newblock {\em IEEE Transactions on Biomedical Engineering}, 44(9):891--900,
  1997.

\bibitem{neuronelimination}
Y.~H. Hu, Q.~Xue, and W.~J. Tompkins.
\newblock Structural simplification of a feed-forward, multi-layer perceptron
  artificial neural network.
\newblock In {\em International Conference on Acoustics, Speech, and Signal
  Processing}, pages 1061--1064, 1991.

\bibitem{moehinton}
R.~A. Jacobs, M.~I. Jordan, S.~J. Nowlan, and G.~E. Hinton.
\newblock Adaptive mixtures of local experts.
\newblock {\em Neural Computation}, 3(1):79--87, 1991.

\bibitem{lowrankexpansion}
M.~Jaderberg, A.~Vedaldi, and A.~Zisserman.
\newblock Speeding up convolutional neural networks with low rank expansions.
\newblock {\em arXiv preprint arXiv:1405.3866}, 2014.

\bibitem{moejacob}
M.~I. Jordan and R.~A. Jacobs.
\newblock Hierarchical mixtures of experts and the {EM} algorithm.
\newblock {\em Neural Computation}, 6(2):181--214, 1994.

\bibitem{tpu}
N.~P. Jouppi, C.~Young, N.~Patil, D.~Patterson, G.~Agrawal, R.~Bajwa, S.~Bates,
  S.~Bhatia, N.~Boden, A.~Borchers, et~al.
\newblock In-datacenter performance analysis of a tensor processing unit.
\newblock In {\em Proceedings of International Symposium on Computer
  Architecture}, pages 1--12, 2017.

\bibitem{lenet5}
Y.~LeCun, L.~Bottou, Y.~Bengio, and P.~Haffner.
\newblock Gradient-based learning applied to document recognition.
\newblock {\em Proceedings of the IEEE}, 86(11):2278--2324, 1998.

\bibitem{transferlearning}
S.~J. Pan and Q.~Yang.
\newblock A survey on transfer learning.
\newblock {\em IEEE Transactions on Knowledge and Data Engineering},
  22(10):1345--1359, 2010.

\bibitem{moeinfinite}
C.~E. Rasmussen and Z.~Ghahramani.
\newblock Infinite mixtures of gaussian process experts.
\newblock In {\em Advances in Neural Information Processing Systems}, pages
  881--888, 2002.

\bibitem{moegoogle}
N.~Shazeer, A.~Mirhoseini, K.~Maziarz, A.~Davis, Q.~Le, G.~Hinton, and J.~Dean.
\newblock Outrageously large neural networks: The sparsely-gated
  mixture-of-experts layer.
\newblock {\em arXiv preprint arXiv:1701.06538}, 2017.

\bibitem{deeplanguagetrans}
I.~Sutskever, O.~Vinyals, and Q.~V. Le.
\newblock Sequence to sequence learning with neural networks.
\newblock In {\em Advances in Neural Information Processing Systems}, pages
  3104--3112, 2014.

\bibitem{moegaussian}
V.~Tresp.
\newblock Mixtures of gaussian processes.
\newblock In {\em Advances in Neural Information Processing Systems}, pages
  654--660, 2001.

\bibitem{transferablefeature}
J.~Yosinski, J.~Clune, Y.~Bengio, and H.~Lipson.
\newblock How transferable are features in deep neural networks?
\newblock In {\em Advances in Neural Information Processing Systems}, pages
  3320--3328, 2014.

\bibitem{ourpaper}
B.~Zhang, A.~Davoodi, and Y.-H. Hu.
\newblock Exploring energy and accuracy tradeoff in structure simplification of
  trained deep neural networks.
\newblock In {\em IEEE Asia and South Pacific Design Automation Conference},
  pages 331--336, 2018.

\end{thebibliography}

\end{document}